\documentclass[10pt,journal]{IEEEtran}
\usepackage{amssymb,amsthm,amsmath,graphicx,cite,float,multirow,array,bm}
\usepackage[ruled]{algorithm2e}

\makeatletter

\newcommand{\Rmnum}[1]{\expandafter\@slowromancap\romannumeral #1@}
\makeatother

\begin{document}
\title{Normalized Total Gradient: A New Measure for Multispectral Image Registration}
\author{Shu-Jie Chen and Hui-Liang~Shen,~\IEEEmembership{Member,~IEEE}

\thanks{This work was supported by the National Natural Science Foundation of China under Grant 61371160, and in part by the Hong Kong Research Institute of Textiles and Apparel (HKRITA) under Grant ITP/048/13TP. \emph{(Corresponding author: Hui-Liang Shen.)}}
\IEEEcompsocitemizethanks{\IEEEcompsocthanksitem The authors are with the College of Information Science and Electronic Engineering, Zhejiang University, Hangzhou 310027, China (e-mail: incredible77@zju.edu.cn; shenhl@zju.edu.cn).}}

\IEEEcompsoctitleabstractindextext{
\begin{abstract}
Image registration is a fundamental issue in multispectral image processing. In filter wheel based multispectral imaging systems, the non-coplanar placement of the filters always causes the misalignment of multiple channel images. The selective characteristic of spectral response in multispectral imaging raises two challenges to image registration. First, the intensity levels of a local region may be different in individual channel images. Second, the local intensity may vary rapidly in some channel images while keeps stationary in others. Conventional multimodal measures, such as mutual information, correlation coefficient, and correlation ratio, can register images with different regional intensity levels, but will fail in the circumstance of severe local intensity variation. In this paper, a new measure, namely normalized total gradient (NTG), is proposed for multispectral image registration. The NTG is applied on the difference between two channel images. This measure is based on the key assumption (observation) that the gradient of difference image between two aligned channel images is sparser than that between two misaligned ones. A registration framework, which incorporates image pyramid and global/local optimization, is further introduced for rigid transform. Experimental results validate that the proposed method is effective for multispectral image registration and performs better than conventional methods.
\end{abstract}

\begin{IEEEkeywords}
Multispectral image, multimodel image, filter wheel,  image registration, image alignment, similarity measure, total gradient, sparsity, rigid transform, affine transform, global optimization, local optimization, intensity variation.
\end{IEEEkeywords}
}

\maketitle
\IEEEdisplaynotcompsoctitleabstractindextext
\IEEEpeerreviewmaketitle

%%%%%%%%% BODY TEXT
\section{Introduction}\label{sec:intro}
Multispectral color imaging has attracted intensive interest in recent years as it can acquire more spectral information than traditional RGB cameras. A multispectral imaging system can be set up by using a camera and an optical device that splits the visible spectrum reflected from the imaged object. Typical light-splitting devices include electronically controlled tunable filters \cite{hardeberg2002multispectral,katravsnik2013} or mechanically controlled filter wheel \cite{brauers2011}. Compare with tunable filters, the filter wheel has the flexibility in filter selection and can achieve high spectral transmittance in the whole visible spectrum range. These two attributes are essential to general-purpose high-quality multispectral imaging.

\begin{figure}[tb]
\centering
\includegraphics[scale=0.9]{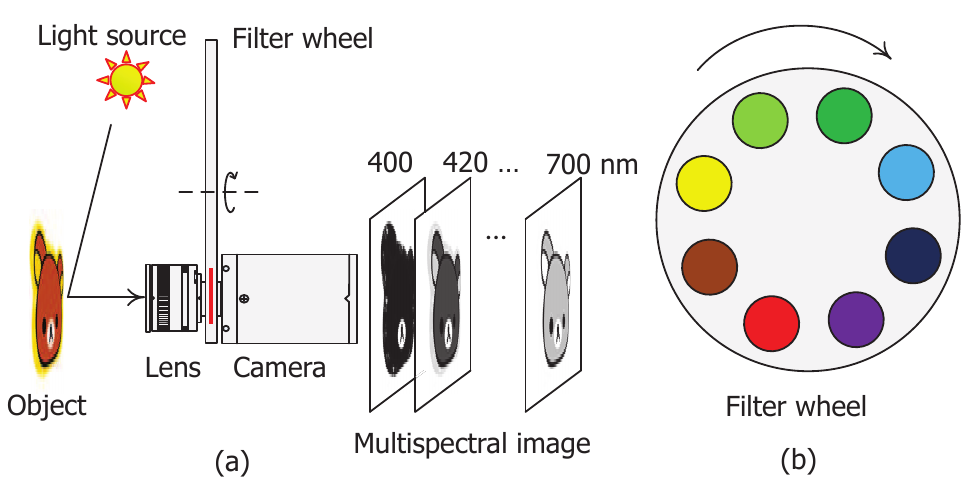}
\caption{The multispectral imaging system (a) and filter wheel (b) used in this work. Note that in our system totally $16$  filters are installed on the wheel, but only $8$ filters are shown for illustration purpose.} \label{fig:Camera}
\end{figure}

In this work, the multispectral images are acquired using a filter wheel based  imaging system. As illustrated in Fig.~\ref{fig:Camera}, the system consists of a monochrome digital camera and a filter wheel installed with a series of optical narrowband filters. These filters are of central wavelengths $400$, $420$, ..., $700$ nm and full width at half maximum (FWHM) $10$ nm. The channel images are sequentially captured by positioning the filters between the lens and camera. Note that for illustration purpose only $8$ filters are shown in Fig.~\ref{fig:Camera}.

The filter wheel based multispectral imaging system usually suffers from image misalignment and out-of-focus blur. The misalignment between channel images is caused by the non-coplanar placement of filters \cite{brauers2011}. It severely precludes the direct use of multispectral images. As shown in Fig.~\ref{fig:multispectral}, the misalignment causes chromatic abberation in the multispectral image (displayed in RGB). Though calibration-based technique for multispectral image alignment is available \cite{brauers2011}, the system must be re-calibrated when the imaging distance changes. Hence, it is more convenient and favored to explore a calibration-free multispectral image registration method for practical applications. On the other hand, the out-of-focus blur is caused by the different effective focal lengths at individual channels, which originates from the fact that the refractive indices of the lens are wavelength dependent \cite{chen2015multispectral}. To deal with this problem, an autofocus method \cite{shen2012autofocus} has been proposed to compute the focus positions of lens using a focusing device (step motor). Alternatively, when the channel images are well aligned, the blur can also be computationally removed by multispectral image deblurring \cite{chen2015multispectral}. In this work, we will show in Section \ref{sec:app} that multispectral images can be restored based on the proposed registration framework and the deblurring method introduced in \cite{chen2015multispectral}. In this way, neither the calibration board nor the focusing device is further needed in the multispectral imaging systems.

Image registration aims to find correspondences between two images by maximizing the defined similarity measures. Extensive surveys of registration techniques can be found in \cite{maintz1998,zitova2003}. The techniques can be classified into two categories according to the descriptors used to formulate the similarity measures \cite{zitova2003}, i.e., the feature-based one and intensity-based one. The feature-based methods usually involve feature detection, feature matching, transform model estimation, and resampling. Generally these methods are more efficient but less accurate compared to the intensity-based methods. As one of the most successful feature-based methods, scale invariant feature transform (SIFT) has been widely used in a variety of matching tasks in computer vision\cite{lowe2004}. The intensity-based registration methods have been developed and applied in medical imaging \cite{maintz1998}. Similarity measures, including sum-of-squared-differences (SSD), correlation coefficient (CC), correlation ratio (CR), and mutual information (MI), have been widely employed in image registration. However, these measures are not robust when registering two images with spatially-varying local intensities \cite{myronenko2010intensity}.

\begin{figure}[tb]
\centering
\includegraphics[scale=0.85]{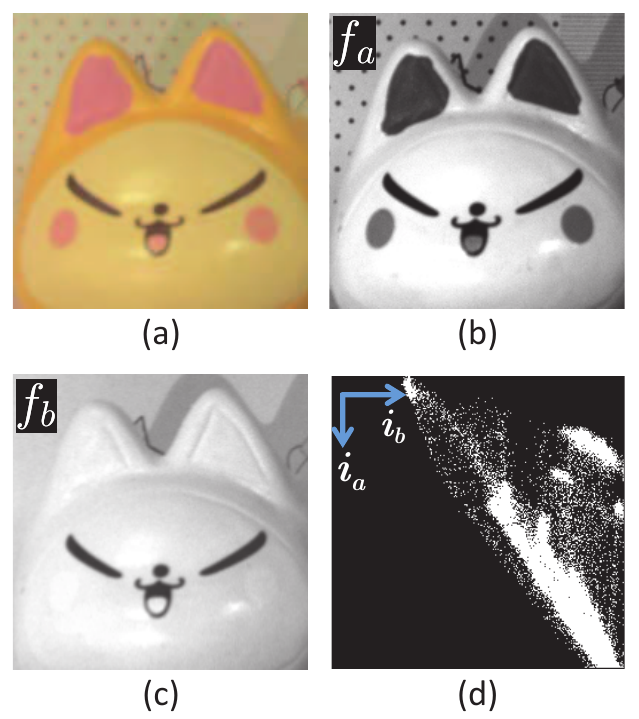}
\caption{Local intensity variation of an aligned multispectral image.  (a) Multispectral image displayed in RGB. (b) Image $f_a$ at channel $9$ ($560$ nm). (c) Image $f_b$ at channel $16$ ($700$ nm). (d) Joint histogram of the two channel images. Notice that the pink spots on the cat head are sensed in image $f_a$ but invisible in image $f_b$.} \label{fig:Joint}
\end{figure}

Multispectral image challenges the conventional intensity-based registration methods at two aspects. First, the intensity levels of the corresponding local region in individual channel images can be essentially different. Second, the local intensity may vary in some channel images while keeps stationary in other channel images. Figure~\ref{fig:Joint} illustrates such an example. The contrast between the pink regions and the background is well sensed in the $9$th channel image $f_a$, but vanishes in the $16$th channel image $f_b$. Consequently, there does not exist a functional mapping between the gray levels $i_a$ and $i_b$ in the joint histogram illustrated in Fig.~\ref{fig:Joint}(d).

\begin{figure}[tb]
  \centering
  \includegraphics[scale=0.78]{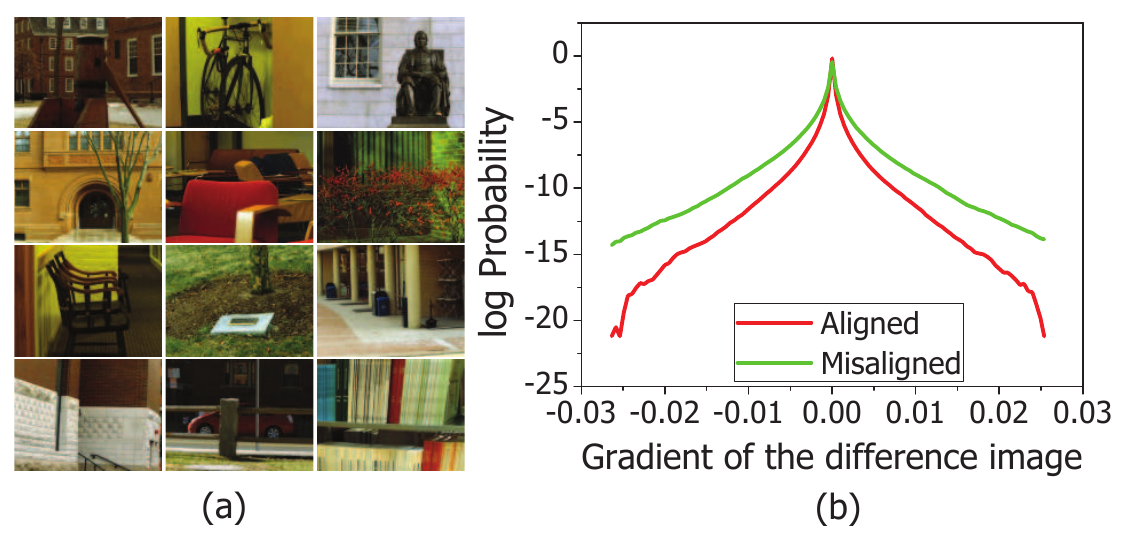}\\
  \caption{Multispectral images of real-world scenes displayed in RGB (a) and distribution of gradient of difference image between channel images (b). }\label{fig:distribution}
\end{figure}

This paper proposes a new intensity-based method, which can resolve the mentioned challenges, for multispectral image registration. The normalized total gradient (NTG), which is applied on the difference between two channel images (referred as \emph{difference image} hereafter), is introduced to measure the alignment of two images. The employment of NTG is based on the key assumption that the gradient of difference image between two aligned images is sparser than that between two misaligned images. This assumption is validated by the statistical investigation on a large number of multispectral images of real-scenes (see Fig.~\ref{fig:distribution}). Based on the NTG, an image registration framework, which incorporates image pyramid and global/local optimization, is introduced for general affine transform. During the minimization, an upside down image pyramid is first constructed for computation acceleration. Differential evolution \cite{price2006}, a powerful global optimization algorithm, is then used to find the initial optimal point at the bottom layer of the pyramid. Finally Newton's method is implemented to successively improve the solution at each layers. Using such a framework, the global optimal affine transform can be found.

The success of the NTG measure in multispectral image registration is due to two reasons. First, the gradient operation weakens the influence of the local intensity that is slowly varying. Second, the sparseness measure allows for rapidly varying local intensity. In fact, the rapid intensity variation contributes rather few additional edges in the gradient map of the difference image, and consequently the sparseness of the gradient map is rarely affected.

To summarize, the main contribution of this work is twofold. First, the NTG is introduced as a registration measure based on the assumption that the gradient of difference image is most sparsely distributed when the two channel images are aligned. Second, a framework, which comprises image pyramid and global/local optimization, is proposed for multispectral image registration. Experimental results show that the proposed method can efficiently register both multispectral and multimodal images.

The remainder of this paper is organized as follows. Section \ref{sec:relatedwork} reviews the related work on multispectral/multimodal image registration. The NTG is introduced in Section \ref{sec:ntg} and the image registration framework is described in Section \ref{sec:framework}. The computation details of optimization is presented in Section \ref{sec:details}. Section \ref{sec:experiment} shows the experimental results and Section \ref{sec:app} illustrates the applications of the proposed image registration method. Finally, Section \ref{sec:conclusion} concludes the paper.

%------------------------------------------------------------------------

\section{Related Work}\label{sec:relatedwork}
Intensity-based image registration can be further divided into unimodal and multimodal image registration based on the criterion whether corresponding pixels (voxels) have similar intensity values. It is clear that multispectral image registration belongs to the latter one. In the following only the most relevant registration methods are reviewed.

\subsection{Mutual Information}
Mutual information (MI) was first introduced for medical image registration in \cite{maes1997} and was extensively investigated in \cite{pluim2003}. MI is defined as
\begin{equation}
I(A,B)=H(A)+H(B)-H(A,B),
\end{equation}
 where $H(\cdot)$ denotes the entropy of the corresponding image, and $H(A,B)$ denotes the entropy of the joint histogram of images $A$ and $B$. When two images are correctly registered, corresponding regions should overlap and the joint histogram will show certain clusters for the intensities of those regions. This consequently results in a least joint entropy $H(A,B)$\cite{pluim2003}. Thus maximizing mutual information can be viewed as registering images $A$ and $B$ such that, in the overlap, the information ($H(A)$ and $H(B)$) provided by the images is large and the local regions are in good match. To further increase immunity against overlap, normalized mutual information has been investigated \cite{studholme1999}.

As a variant of MI, regionalized mutual information was introduced to cope with the local intensity variation\cite{studholme2006deformation}. By assuming that the local intensities are slowly varying across the image, a global mutual information is constructed from local regional distributions and hence the registration accuracy improves. Additionally, the limitation on slowly varying local intensity can also be avoided by computing MI only in the regions with large spatial variations (e.g., corner and edge). This idea is implemented in \cite{woo2014} and \cite{luan2008multimodality} by augmenting MI with the Harris operator or saliency measure. Such methods perform relatively well on images with slow intensity variation at the cost of high computation complexity.

\subsection{Correlation Ratio}
While MI describes the clustering property of joint histogram by entropy, correlation ratio (CR) characterizes such property by variance\cite{roche1998}. CR stems from the law of total variance
\begin{equation}\label{eq:LawofVar}
\text{var}(A)=\mathbb{E}_B(\text{var}(A|B))+\text{var}_B(\mathbb{E}(A|B)).
\end{equation}
When the joint histogram is well clustered, for every single intensity which represents some local region in $B$, the variance of intensity in the same local region of $A$ should be small. This leads to a small value of $\text{var}(A|B)$. For all intensities of $B$ in the region of overlap, the mean variance of $A$ denoted by $\mathbb{E}_B(\text{var}(A|B))$ are hence the smallest when registered. Accordingly, the proportion of the second term of the right hand side of (\ref{eq:LawofVar}) arrives its maximum, which indicates the correlation ratio
\begin{equation}\label{eq:cr}
CR=\dfrac{\text{var}_B(\mathbb{E}(A|B))}{\text{var}(A)}
\end{equation}  is maximized.

Correlation ratio was thrown light upon again in the view of pattern matching recently. The work \cite{hel2014} starts from the discussion of matching by tone mapping (MTM) and ends up with a variant of CR. According to \cite{hel2014}, the pattern to window distance in MTM is computed as
\begin{equation}\label{eq:mtm0}
 D(\mathbf{q},\mathbf{w})=\dfrac{1}{|\mathbf{w}|\; \text{var}(\mathbf{w})}\left( \|\mathbf{w}\|^2-\displaystyle\sum_j{\dfrac{1}{|\mathbf{q}^j|}(\mathbf{q}^j \cdot\mathbf{w})^2}\right),
\end{equation}
where $\mathbf{w}$ is the window patch, $\mathbf{q}^j$ is the $j$th slice in pattern $\mathbf{q}$, $|\mathbf{q}^j|$ is the pixel number in slice $\mathbf{q}^j$, and $|\mathbf{w}|$ is the pixel number of the window patch. Note that $\sum_j|\mathbf{q}^j|=|\mathbf{w}|$. The distance (\ref{eq:mtm0}) can be reformulated as
\begin{equation}\label{eq:mtm}
 \begin{split}
D(\mathbf{q},\mathbf{w})&=\dfrac{1}{|\mathbf{w}|\; \text{var}(\mathbf{w})}\\
&\sum_j{ |\mathbf{q}^j|\left(\dfrac{\sum_{i\in\mathbf{q}^j} w_{ji}^2}{|\mathbf{q}^j|}-\left(\dfrac{\sum_{i\in\mathbf{q}^j} w_{ji}}{|\mathbf{q}^j|}\right)^2 \right)},
\end{split}
\end{equation}
where $w_{ji}$ denotes the intensity of the $i$th pixel in the $j$th slice of the window patch specified by the $j$th slice of the pattern. Equation (\ref{eq:mtm}) can be further transformed to
\begin{equation}\label{eq:mtm1}
D(\mathbf{q},\mathbf{w})=\dfrac{1}{|\mathbf{w}|\; \text{var}(\mathbf{w})}\displaystyle\sum_j |\mathbf{q}^j|\text{var}(\mathbf{w}^j),
\end{equation}
where $\mathbf{w}^j$ is the $j$th slice of the window patch with respect to pattern $\mathbf{q}^j$. With some transformation, it can be validated that the distance (\ref{eq:mtm1}) equals to $(1-CR)$. Hence MTM and CR are actually equivalent.

It should be noted that both MI and CR describe the regional correspondence of two different modal images. When the intensity of two images can be well approximated by functional mapping, which allows for a small intensity variation of a single local region, both MI and CR can achieve outstanding registration performance. MI is more robust than CR with respect to intensity variation because entropy is more accurate in measuring the clustering property than variance. Entropy measures the sparseness of the joint histogram and allows for a wider range of distributions of intensity while variance is only best suited to intensity that follows the normal distribution. However, such functional correspondence does not always exist in multispectral images. Figure~\ref{fig:Joint} shows an example of aligned patch of channel $9$ ($560$ nm) and channel $16$ ($700$ nm). The joint histogram of these two patches cannot be approximated as a functional mapping from the intensity of $f_a$ to that of $f_b$, but the joint distribution still clusters.

\subsection{Local Normalized Correlation Coefficient}
While MI and CR are global intensity-based registration methods, local normalized correlation coefficient (NCC) \cite{irani1998} has been introduced to measure the correspondence between two modal images locally. Based on the assumption that the directional derivative energy maps are locally linearly correlated, normalized cross correlation can be used to measure the similarity between two multimodal patches. However, since not all patches satisfy this assumption, an outlier rejection strategy is needed when constructing a global cost function. To weaken the local linearity assumption, a more robust variant of NCC called robust selective normalized cross correlation (RSNCC) criterion \cite{shen2014} was recently proposed to solve the dense matching problem of both multispectral and natural images. As stated in \cite{shen2014}, though structure inconsistency and strong outliers caused by shadow and highlight are well handled in the registration framework, large errors may occur on regions that do not contain informative features.

The local linearity assumption is a strong functional dependence of descriptors in regions specified by user-defined windows. Such descriptors can be the intensity directly or the gradient. Either explicit outlier rejection strategy or implicit robust selective function aims to reduce the risk of registration inaccuracy, which results from those regions that violate such assumption. However, the procedure of selection is empirical and increases the computation complexity.

\subsection{Residual Complexity}
Our proposed method is closely related to the work in \cite{myronenko2010intensity} in the sense of measuring the complexity of the difference image. In \cite{myronenko2010intensity} the difference of two unimodal images with local intensity variation, referred as the residual image, was used. The author showed that the complexity of the residual image (RC) reduced to its minimum when two images were aligned. RC can be simply represented by using sparse formulation of discrete cosine transform (DCT) coefficients.

However, RC is less suitable for multispectral image registration. When two images are of different modality, the sparseness of the DCT coefficients of the difference image is no longer guaranteed. The reason is that the intensities of two corresponding regions do not counteract while registered. Compared to RC, the proposed method measures the sparseness of the gradient of difference image. By taking the gradient, the residual intensity counteracts with its neighbors so that multi-modality impacts least on the sparseness.

% The residual intensities of different regions therefore increases the sparseness.

%------------------------------------------------------------------------
\section{Normalized Total Gradient (NTG)}\label{sec:ntg}
Statistics of image features are of significant use in image processing and computer vision. The sparseness of gradients of natural images has been prevalently applied in single image deblurring \cite{fergus2006}, denoising \cite{portilla2003}, and inpainting \cite{levin2003}.  Statistics of multispectral images have also been explored for the efficient representation of multispectral images \cite{chakrabarti2011}. In this work, the gradient of difference image is studied and the sparseness of such feature is employed in multispectral image registration.

%-------------------------------------------------------------------------
\subsection{Sparseness}

To validate the sparseness assumption of the gradient of difference image, a multispectral image database with $77$ real-world scenes from \cite{chakrabarti2011} is employed. Some images are shown in Fig.~\ref{fig:distribution}(a). Each multispectral image consists of $31$ channels and covers the visible spectrum from $420$ to $720$ nm at an interval of 10 nm. Without loss of generality, image intensities are normalized to the $[0, 1]$ range. All multispectral images are originally aligned. Random displacements are imposed on all the channels, except for the reference channel, to synthesize the misaligned multispectral images. The $16$th channel is chosen to serve as the reference. The \emph{difference images} are then computed as the differences between the reference and the other channel images.

The distributions of the gradient of difference image in cases of alignment and misalignment are shown in Fig.~\ref{fig:distribution}(b). It is observed that the distribution corresponding to the aligned channel images is more heavy-tailed than that corresponding to the misaligned ones. More zero gradients and less large gradients of the difference image are obtained by registration. In other words, the gradient of difference image becomes sparser when the multispectral image is aligned. This can be mathematically formulated as
\begin{equation}\label{eq:tg}
 \sum_l|\nabla_l \{f-f_R\}| \leq \sum_l|\nabla_l\{\tilde{f}-f_R\}|,
\end{equation}
where $f_R$ represents the reference channel image, $f$ and $\tilde{f}$ denote the aligned and misaligned channel images, respectively. The operator $\nabla_l$, with $l \in \{x, y\}$, denotes the gradient computation along the direction $l$. Notice that $|\cdot|$ is the $L_1$ norm, the absolute gradient of difference image $f-f_R$ can be explicitly expanded as
$$
|\nabla_l\{f-f_R\}| = \sum_\mathbf{x}{|\nabla_l \{f(\mathbf{x})-f_R(\mathbf{x})\}|},
$$
where $\mathbf{x}=(x,y)^\mathsf{T}$ denotes the spatial location.

The explanation of the sparsest distribution of gradient in circumstance of well alignment is straightforward. It is expected that, when two images are aligned, the gradients of these images generally overlap and hence the number of nonzero gradients of the difference image reduces. A rigorous paradigm that characterizes such number should be the $L_0$ norm. In this work we employ the $L_1$ norm instead since it is computationally efficient and performs comparatively to the $L_0$ norm \cite{candes2006}.

In what follows, we refer to the $L_1$ norm of the gradient of the image along all directions ($x$ and $y$ in this work) as the total gradient (TG) of the image, i.e.,
$$
TG(f) =  \sum_l|\nabla_l f|.
$$
Note that this definition is different to total variation (TV) \cite{rudin1992}. With the definition, the expression (\ref{eq:tg}) can be interpreted that the TG of difference image is minimized when two images are registered. Hence TG functions as a simple measure for multispectral image registration.

\subsection{Normalization}
Region of overlap between two images is composed by those pixels whose intensities or features are used to compute the measure.  A measure is of limited use if it is sensitive to the overlap variation. The work \cite{studholme1999} presents an extensive discussion on the overlap, and proposes the normalized mutual information (NMI) to reduce the effect of varying overlap on entropy measure.

The TG of the difference image is sensitive to the overlap because the summation of absolute gradients is defined on the region of overlap. The TG reduces when the region of overlap shrinks. By considering the $L_1$ norm as a kind of energy measure, we propose the normalized total gradient (NTG) of difference image as follows,
\begin{equation}\label{eq:ntg}
NTG(f,f_R)=\dfrac{\sum_l|\nabla_l\{f-f_R\}|}{\sum_l|\nabla_lf|+\sum_l|\nabla_lf_R|}.
\end{equation}
In (\ref{eq:ntg}), the numerator is the TG of the difference image and the denominator is the total energy that normalizes the TG measure. It can be easily verified that $0\leq NTG \leq 1$. Minimizing NTG is equivalent to registering two images while increasing their total energy in the region of overlap.

\section{Image Registration Using NTG}\label{sec:framework}
The proposed measure NTG is minimized when two images are registered. Though NTG can be applied to a variety of registration tasks, we narrow down to image registration with global affine transform model. The affine transform model is frequently used as the pre-processing of medical image registration\cite{rueckert1999nonrigid}. Besides, it has been established that the multispectral imaging registration in the filter wheel based imaging system can be well approximated by the affine transform \cite{klein2012multispectral}.

\subsection{Affine Transform}
The aim of image registration is to find a transform $\mathbf{p}$ such that the transformed floating image $g(\mathbf{x,p})=f(u(\mathbf{x,p}),v(\mathbf{x,p}))$ matches the reference image $f_R(\mathbf{x})$. The affine transform can be formulated as
\begin{equation}\label{eq:affine}
\begin{pmatrix} u \\ v  \end{pmatrix} = \begin{pmatrix} p_1 &p_2 &p_3 \\ p_4 &p_5 &p_6 \end{pmatrix}\begin{pmatrix} x\\ y \\1\end{pmatrix} = \mathbf{P} \begin{pmatrix} x\\ y\\1 \end{pmatrix},
\end{equation}
where $\mathbf{p}=(p_1,p_2,\cdots,p_6)^\mathsf{T}$ is the vector form of $\mathbf{P}$. According to the discussion above we know that when the transformed floating image $g$ and the reference image $f_R$ are well matched, the $NTG(g,f_R)$ measure is minimized. Hence registering the floating image $f$ and the reference image $f_R$ is equivalent to solving the following optimization problem,
\begin{equation}\label{eq:obj}
\begin{aligned}
\min_{\mathbf{p}} J(\mathbf{p})&\equiv NTG(g,f_R)\\
&=\dfrac{\sum_l\sum_{\mathbf{x}\in\Omega(\mathbf{p})}|\nabla_l\{g(\mathbf{x},\mathbf{p})-f_R(\mathbf{x})\}|}
{\sum_l\sum_{\mathbf{x}\in\Omega(\mathbf{p})}\left(|\nabla_lg(\mathbf{x},\mathbf{p})|+|\nabla_lf_R(\mathbf{x})|\right)},
\end{aligned}
\end{equation}
where $g(\mathbf{x},\mathbf{p}) = f(u(\mathbf{x,p}),v(\mathbf{x,p})), ~(u,v)^\mathsf{T}=\mathbf{P}(x,y,1)^\mathsf{T}$, $l\in\{x,y\}$, and $\Omega(\mathbf{p})$ denotes the region of overlap.

\subsection{Image Registration Framework}
While the problem specified in (\ref{eq:obj}) is a global optimization problem, local optimization methods can result in a transform that corresponds to a local minimum of the objective function, despite their simple and fast computation. On the other hand, global optimization methods are not advisable to be directly employed due to their unacceptable computation time when images are large. Hence we employ a hybrid strategy to balance computation efficiency and accuracy.

Images are first subsampled into $K$ layers to form an upside down image pyramid whose lower layer is the subsampled version of the upper one. Then global optimization is imposed on the bottom layer to furnish a good initial estimate for the successive optimization. The initial point is transferred to the upper layer and is refined by local optimization until the top layer is reached. In this work the differential evolution \cite{price2006} and the well-known Newton's method are used as the global and local optimizers, respectively.

Differential evolution (DE) is a genuinely useful global optimization algorithm and  has earned a reputation as a very effective and reliable global optimizer. Similar to most evolution algorithms, three steps are needed after choosing the initial population, namely, mutation, crossover, and selection. The initial vector population is chosen randomly and covers the constrained parameter space. New parameter vectors are generated by adding the weighted difference between two population vectors to a third one. The mutated vector's parameters are then mixed with the parameters of another predetermined vector (the target vector), yielding the trial vector. The target vector is replaced if the trial vector has a lower cost value. This has been done over all population vectors so that competitions take place in each generation \cite{storn1997}.

In the global optimization, we use the DeMat library \cite{price2006} to search the initial point at the bottom layer. A population of $30$ individuals and $200$ generations are usually adequate for deriving a good initial point. With a four-layered image pyramid, the searching space of $\mathbf{p}=(p_1,\cdots,p_6)^\mathsf{T}$ is constrained from $[0.95, -0.05, -10, -0.05, 0.95, -10]$ to $[1.05, 0.05, 10, 0.05, 1.05, 10]$ since no severe scaling and rotation take place in the captured multispectral images. Other DE parameters are set by default\cite{price2006}.

In the local optimization, the parameters at each layer is updated as
\begin{equation}
\mathbf{p}^{t+1} = \mathbf{p}^t-(\mathbf{H}_{\mathbf{p}}^J)^{-1}J_{\mathbf{p}}|_{\mathbf{p=p}^t},
\end{equation}
where $\mathbf{H}_{\mathbf{p}}^J$ and $J_\mathbf{p}$ are, respectively, the Hessian matrix and gradient of the objective function $J(\mathbf{p})$ with respect to $\mathbf{p}$. The superscribe $t$ represents the $t$th iteration. When the initial point locates near the optimal solution, which is guaranteed by global optimization at the bottom layer, $6$ iterations are usually sufficient for the convergence of Newton's method \cite{wright1999}.

\section{Implementation Details of Newton's Method}\label{sec:details}
To minimize the objective function $J(\mathbf{p})$ in (\ref{eq:obj}) we first denote $m(\mathbf{p})$ and $n(\mathbf{p})$ as the numerator and denominator of $J(\mathbf{p})$, respectively, as follows,
$$
m(\mathbf{p}) = \sum_{\mathbf{x}\in \Omega(\mathbf{p})}(|g_x(\mathbf{x,p})-f_{R,x}(\mathbf{x})|+|g_y(\mathbf{x,p})-f_{R,y}(\mathbf{x})|),
$$
$$
n(\mathbf{p}) = \sum_{\mathbf{u}\in \Omega(\mathbf{p})}(|g_x(\mathbf{x,p})|+|g_y(\mathbf{x,p})|+|f_{R,x}(\mathbf{x})|+|f_{R,y}(\mathbf{x})|),
$$
where the subscripts $x$ and $y$ denote the partial derivatives along the $x$ and $y$ directions. With these notations, the objective function (\ref{eq:obj}) is represented as $J(\mathbf{p})=\frac{m(\mathbf{p})}{n(\mathbf{p})}$. By taking the derivative with respect to $\mathbf{p}$ we have
\begin{equation}\label{eq:J_p}
J_\mathbf{p}=n^{-2}(nm_\mathbf{p}-mn_\mathbf{p}),
\end{equation}
where $m_\mathbf{p}$ and $n_\mathbf{p}$ denote the gradient of $m$ and $n$ with respect to $\mathbf{p}$. Note that in (\ref{eq:J_p}) the parameter $\mathbf{p}$ in parenthesis is omitted for notation simplification. The $(i,j)\text{th}$ entry in Hessian matrix $\mathbf{H}_\mathbf{p}^J$ is
$$
\begin{aligned}
\mathbf{H}_\mathbf{p}^J(i,j)
&=J_{p_i,p_j}\\
&=n^{-2}(nm_{p_i,p_j} - mn_{p_i,p_j} + 2mn^{-1}n_{p_i}n_{p_j}\\
&\quad -m_{p_i}n_{p_j}-n_{p_i}m_{p_j}).
\end{aligned}
$$
Hence the Hessian matrix $\mathbf{H}_\mathbf{p}^J$ becomes
\begin{equation}\label{eq:H_p_J}
\begin{aligned}
\mathbf{H}_{\mathbf{p}}^J=
&n^{-2}(n\mathbf{H}_{\mathbf{p}}^m-m\mathbf{H}_\mathbf{p}^n + 2mn^{-1}n_{\mathbf{p}}n_{\mathbf{p}}^\mathsf{T} \\
&-m_{\mathbf{p}}n_{\mathbf{p}}^\mathsf{T} -n_{\mathbf{p}}m_{\mathbf{p}}^\mathsf{T}),
\end{aligned}
\end{equation}
where $\mathbf{H}_{\mathbf{p}}^m$ and $\mathbf{H}_{\mathbf{p}}^n$ denote the Hessian matrices of $m$ and $n$ with respect to $\mathbf{p}$.

To this end, in order to derive the gradient $J_{\mathbf{p}}$ and the Hessian matrix $\mathbf{H}_{\mathbf{p}}^J$ we resort to computing $m_{\mathbf{p}}$, $n_{\mathbf{p}}$, $\mathbf{H}_{\mathbf{p}}^m$, and $\mathbf{H}_{\mathbf{p}}^n$, which will be elaborated below.

\subsection{Computing $m_\mathbf{p}$ and $n_\mathbf{p}$}
Let $d(\mathbf{x,p})$ denote the difference image, i.e.
$$
d(\mathbf{x,p}) = g(\mathbf{x,p})-f_R(\mathbf{x}),
$$
thus it is easy to derive that
$$
d_x=g_x-f_{R,x}, ~d_y=g_y-f_{R,y},
$$
\begin{equation}\label{eq:gradeq}
d_{x,\mathbf{p}} = g_{x,\mathbf{p}}, ~d_{y,\mathbf{p}} = g_{y,\mathbf{p}}.
\end{equation}
Let $\rho_1(\cdot)$ denote the first order derivative of the absolute function $|\cdot|$ (see Subsection \emph{C}). Since the overlap changes slightly in the two consecutive updates of $\mathbf{p}$, i.e. $\Omega_{\mathbf{p}}(\mathbf{p}^t)\approx \mathbf{0}$, by taking the derivatives of $m$ and $n$ with respect to $\mathbf{p}$ we have
\begin{equation}\label{eq:m_p}
m_\mathbf{p}\approx \sum_{\mathbf{x}\in \Omega(\mathbf{p})}(\rho_1(d_x)g_{x,\mathbf{p}}+ \rho_1(d_y)g_{y,\mathbf{p}})
\end{equation}
and
\begin{equation}\label{eq:n_p}
n_\mathbf{p}\approx \sum_{\mathbf{x}\in \Omega(\mathbf{p})}(\rho_1(g_x)g_{x,\mathbf{p}}+ \rho_1(g_y)g_{y,\mathbf{p}}).
\end{equation}

We narrow our focus down to the computation of $g_{x,\mathbf{p}}$ and $g_{y,\mathbf{p}}$ since all the other terms are known given the current transform parameter $\mathbf{p}^t$. By implementing $g(\mathbf{x,p})=f(u(\mathbf{x,p}),v(\mathbf{x,p}))$, we have
$$
\begin{pmatrix}g_x\\g_y\end{pmatrix}=\begin{pmatrix}u_x,v_x\\u_y,v_y\end{pmatrix}\begin{pmatrix}f_u\\f_v\end{pmatrix}.
$$
Consequently it can be derived that
\begin{equation}\label{eq:g_p}
\begin{aligned}
\begin{pmatrix}g_{x,p_i}\\g_{y,p_i}\end{pmatrix}
=&\begin{pmatrix}u_x,v_x\\u_y,v_y\end{pmatrix}\begin{pmatrix}f_{uu},f_{uv}\\f_{vu},f_{vv}\end{pmatrix}\begin{pmatrix}u_{p_i}\\v_{p_i}\end{pmatrix}\\
&+\begin{pmatrix}u_{x,p_i},v_{x,p_i}\\u_{y,p_i},v_{y,p_i}\end{pmatrix}\begin{pmatrix}f_u\\f_v\end{pmatrix},
\end{aligned}
\end{equation}
where $p_i\in\{p_1,p_2,\cdots,p_6\}.$ In (\ref{eq:g_p}), the gradient of the objective function is finally decomposed to the transformed first order and second order gradients of the floating image, and the gradients of the transform model.

According to the affine transform model (\ref{eq:affine}), we have
$$
\begin{pmatrix}u_x,v_x\\u_y,v_y\end{pmatrix}=\begin{pmatrix}p_1,p_4\\p_2,p_5 \end{pmatrix},
$$
$$
\begin{aligned}
&u_\mathbf{p}=(x,y,1,0,0,0)^\mathsf{T},~v_\mathbf{p}=(0,0,0,x,y,1)^\mathsf{T},\\
&u_{x,\mathbf{p}}=(1,0,0,0,0,0)^\mathsf{T},~u_{y,\mathbf{p}}=(0,1,0,0,0,0)^\mathsf{T},\\
&v_{x,\mathbf{p}}=(0,0,0,1,0,0)^\mathsf{T},~v_{y,\mathbf{p}}=(0,0,0,0,1,0)^\mathsf{T}.
\end{aligned}
$$
To this end, the gradient of the objective function can be computed by inserting (\ref{eq:m_p}), (\ref{eq:n_p}), and (\ref{eq:g_p}) into (\ref{eq:J_p}).

\begin{figure}[t]
  \centering
  \includegraphics[scale = 0.9]{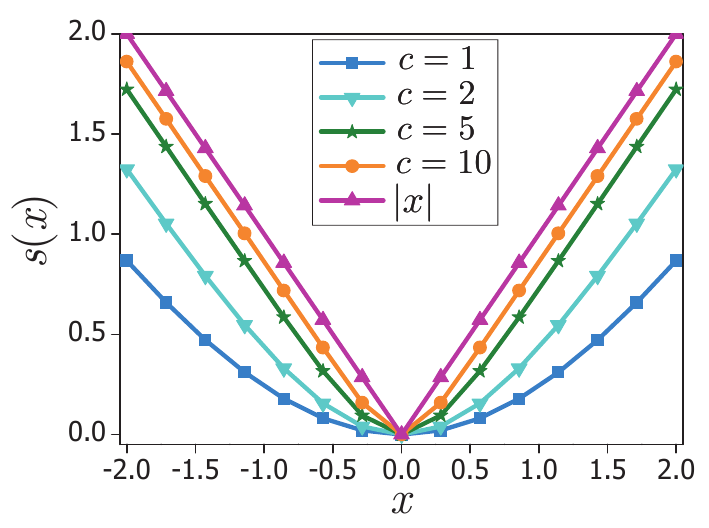}\\
  \caption{Function $s(x)$ with respect to different values of $c$, The value of $x$ is in the range $[-2, 2]$. The absolute function $|x|$ can be well approximated by $s(x)$ when $c=10$.}\label{fig:abs}
\end{figure}

\subsection{Computing $\mathbf{H}^m_\mathbf{p}$ and $\mathbf{H}^n_\mathbf{p}$}
 Let $\rho_2(\cdot)$ be the second order derivative of the absolute function $|\cdot|$. By using (\ref{eq:gradeq}) and (\ref{eq:m_p}), the $(i,j)$th entry in $\mathbf{H}^m_\mathbf{p}$ can be computed as
$$
\begin{aligned}
\mathbf{H}^m_\mathbf{p}(i,j)&=m_{p_i,p_j}\\
&\approx \sum_{\mathbf{x\in \Omega(p)}}(\rho_1(d_x)g_{x,p_i,p_j}+\rho_2(d_x)g_{x,p_i}g_{x,p_j} \\
&\qquad\qquad + \rho_1(d_y)g_{y,p_i,p_j}+\rho_2(d_y)g_{y,p_i}g_{y,p_j}).
\end{aligned}
$$
Hence the Hessian matrix $\mathbf{H}^m_\mathbf{p}$ becomes
\begin{equation}\label{eq:H_p_m}
\begin{aligned}
\mathbf{H}_{\mathbf{p}}^m\approx\sum_{\mathbf{x\in \Omega(p)}}(
&\rho_1(d_x)\mathbf{H}_\mathbf{p}^{g_x}+ \rho_2(d_x)g_{x,\mathbf{p}}g_{x,\mathbf{p}}^\mathsf{T}\\
&\rho_1(d_y)\mathbf{H}_\mathbf{p}^{g_y}+ \rho_2(d_y)g_{y,\mathbf{p}}g_{y,\mathbf{p}}^\mathsf{T}),
\end{aligned}
\end{equation}
where $\mathbf{H}_\mathbf{p}^{g_x}$ and $\mathbf{H}_\mathbf{p}^{g_y}$ are the Hessian matrices of $g_x$ and $g_y$.

Similarly, we have
$$
\begin{aligned}
\mathbf{H}^n_\mathbf{p}(i,j)&=n_{p_i,p_j}\\
&\approx \sum_{\mathbf{x\in \Omega(p)}}(\rho_1(g_x)g_{x,p_i,p_j}+\rho_2(g_x)g_{x,p_i}g_{x,p_j} \\
&\qquad\qquad + \rho_1(g_y)g_{y,p_i,p_j}+\rho_2(g_y)g_{y,p_i}g_{y,p_j})
,
\end{aligned}
$$
and thus
\begin{equation}\label{eq:H_p_n}
\begin{aligned}
\mathbf{H}_{\mathbf{p}}^n\approx\sum_{\mathbf{x\in \Omega(p)}}(
&\rho_1(g_x)\mathbf{H}_\mathbf{p}^{g_x}+ \rho_2(g_x)g_{x,\mathbf{p}}g_{x,\mathbf{p}}^\mathsf{T}\\
&\rho_1(g_y)\mathbf{H}_\mathbf{p}^{g_y}+ \rho_2(g_y)g_{y,\mathbf{p}}g_{y,\mathbf{p}}^\mathsf{T}).
\end{aligned}
\end{equation}

According to the affine model given in (\ref{eq:g_p}), we can derive that
\begin{equation}\label{eq:g_pq}
\begin{aligned}
\begin{pmatrix}g_{x,p_i,p_j}\\g_{y,p_i,p_j}\end{pmatrix}=
&\begin{pmatrix}u_{x,p_j},v_{x,p_j}\\u_{y,p_j},v_{y,p_j}\end{pmatrix}\begin{pmatrix}f_{uu},f_{uv}\\f_{vu},f_{vv}\end{pmatrix}\begin{pmatrix}u_{p_i}\\v_{p_i}\end{pmatrix}\\
&+\begin{pmatrix}u_x,v_x\\u_y,v_y\end{pmatrix}\begin{pmatrix}f_{uu,p_j},f_{uv,p_j}\\f_{vu,p_j},f_{vv,p_j}\end{pmatrix}\begin{pmatrix}u_{p_i}\\v_{p_i}\end{pmatrix}\\
&+\begin{pmatrix}u_{x,p_i},v_{x,p_i}\\u_{y,p_i},v_{y,p_i}\end{pmatrix}\begin{pmatrix}f_{u,p_j}\\f_{v,p_j}\end{pmatrix},
\end{aligned}
\end{equation}
where
$$
\begin{pmatrix}f_{u,p_j}\\f_{v,p_j}\end{pmatrix}=\begin{pmatrix}f_{uu},f_{uv}\\f_{vu},f_{vv}\end{pmatrix}\begin{pmatrix}u_{p_j}\\v_{p_j}\end{pmatrix},
$$
$$
\begin{pmatrix}f_{uu,p_j}\\f_{vu,p_j}\end{pmatrix}=\begin{pmatrix}f_{uuu},f_{uuv}\\f_{vuu},f_{vuv}\end{pmatrix}\begin{pmatrix}u_{p_j}\\v_{p_j}\end{pmatrix},
$$
and
$$
\begin{pmatrix}f_{uv,p_j}\\f_{vv,p_j}\end{pmatrix}=\begin{pmatrix}f_{uvu},f_{uvv}\\f_{vvu},f_{vvv}\end{pmatrix}\begin{pmatrix}u_{p_j}\\v_{p_j}\end{pmatrix}.
$$
It is clear that the Hessian matrix of the objective function can finally decomposed to the transformed gradients of the floating image up to third orders. To this end, the Hessian matrix can be computed by inserting (\ref{eq:H_p_m}), (\ref{eq:H_p_n}), and (\ref{eq:g_pq}) into (\ref{eq:H_p_J}).

\subsection{Form of $\rho_1$ and $\rho_2$}\label{sec:abs}
Since the absolute function $|x|$ is not differentiable at $x=0$, we use the function
\begin{equation}
s(x) = x + \dfrac{2}{c}\log(\dfrac{1+e^{-cx}}{2})
\end{equation}
to approximate it. Figure~\ref{fig:abs} illustrates the function $s(x)$ with respect to different values of $c$. The value of $x$, which is the gradient of difference image, is in the range $[-2, 2]$. As illustrated, $s(x)$ gradually approximates $|x|$ as $c$ increases. In this work we set $c=10$. The first and second derivatives of $s(x)$ are, respectively,
\begin{equation}
\rho_1(x) = \dfrac{1-e^{-cx}}{1+e^{-cx}}
\end{equation}
and
\begin{equation}
\rho_2(x)=\dfrac{2c}{e^{-cx}+e^{cx}+2}.
\end{equation}

\begin{figure}[tb]
  \centering
  \includegraphics[scale=0.65]{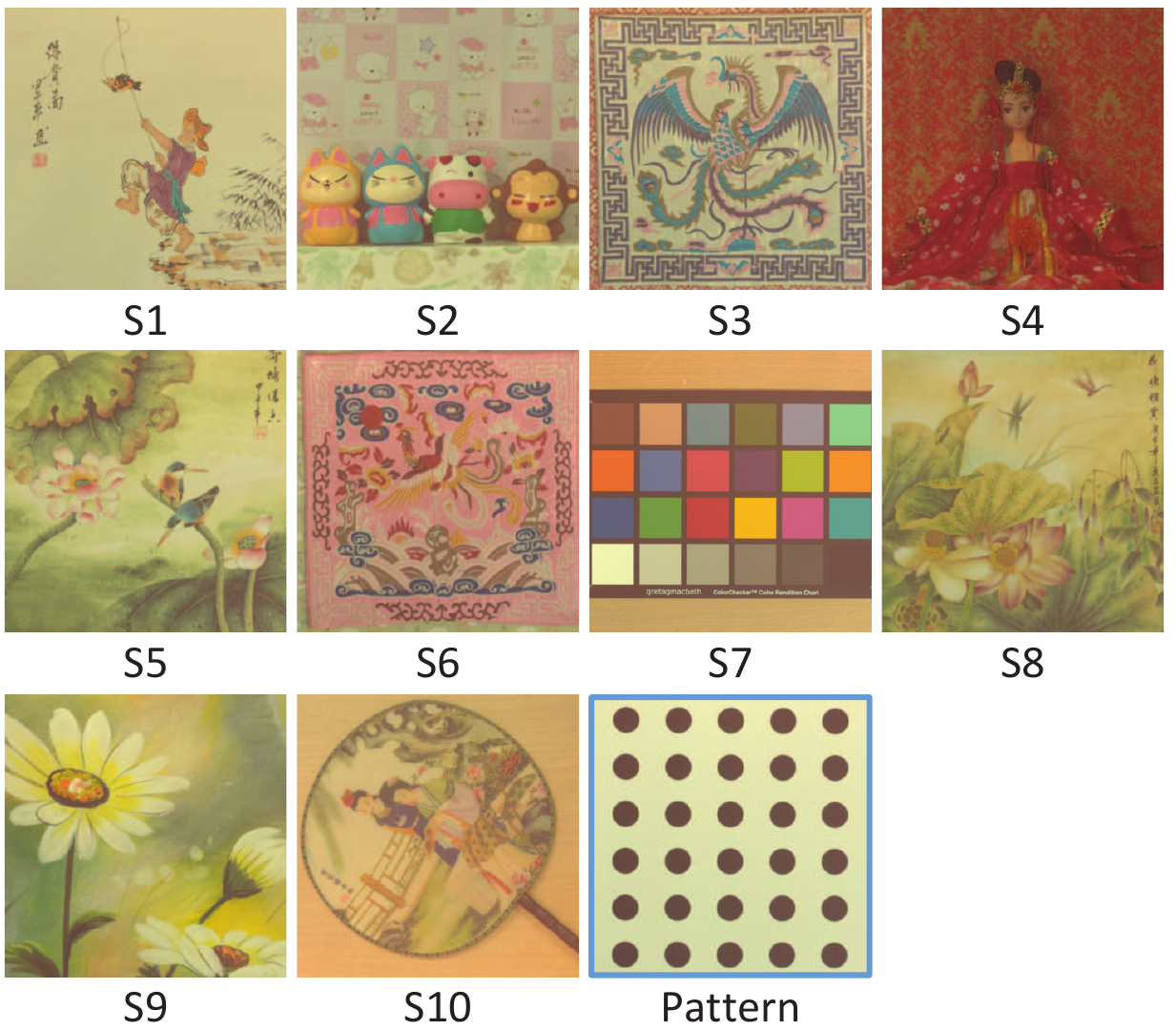}\\
  \caption{The multispectral sample images (\emph{S1}, \emph{S2}, $\cdots$, \emph{S10}) and pattern image (displayed in RGB) used in the experiments. The pattern image is employed in the quantitative evaluation of registration accuracy. }\label{fig:database}
\end{figure}

\section{Experiments}\label{sec:experiment}
Experiments were conducted on $10$ multispectral images of real-scenes which are shown in Fig. \ref{fig:database}. These images were captured using the imaging system illustrated in Fig. \ref{fig:Camera}. The performances of MI\cite{maes1997}, RC\cite{myronenko2010intensity}, CR\cite{roche1998}, RSNCC\cite{shen2014} and the proposed NTG were compared at two aspects. First, the robustness of these measures was investigated on multispectral images with local intensity variation. Second, the registration accuracy was quantitatively evaluated on real multispectral images.

\subsection{Robustness to Intensity Variation}
\begin{figure*}
\centering
\includegraphics[scale=0.9]{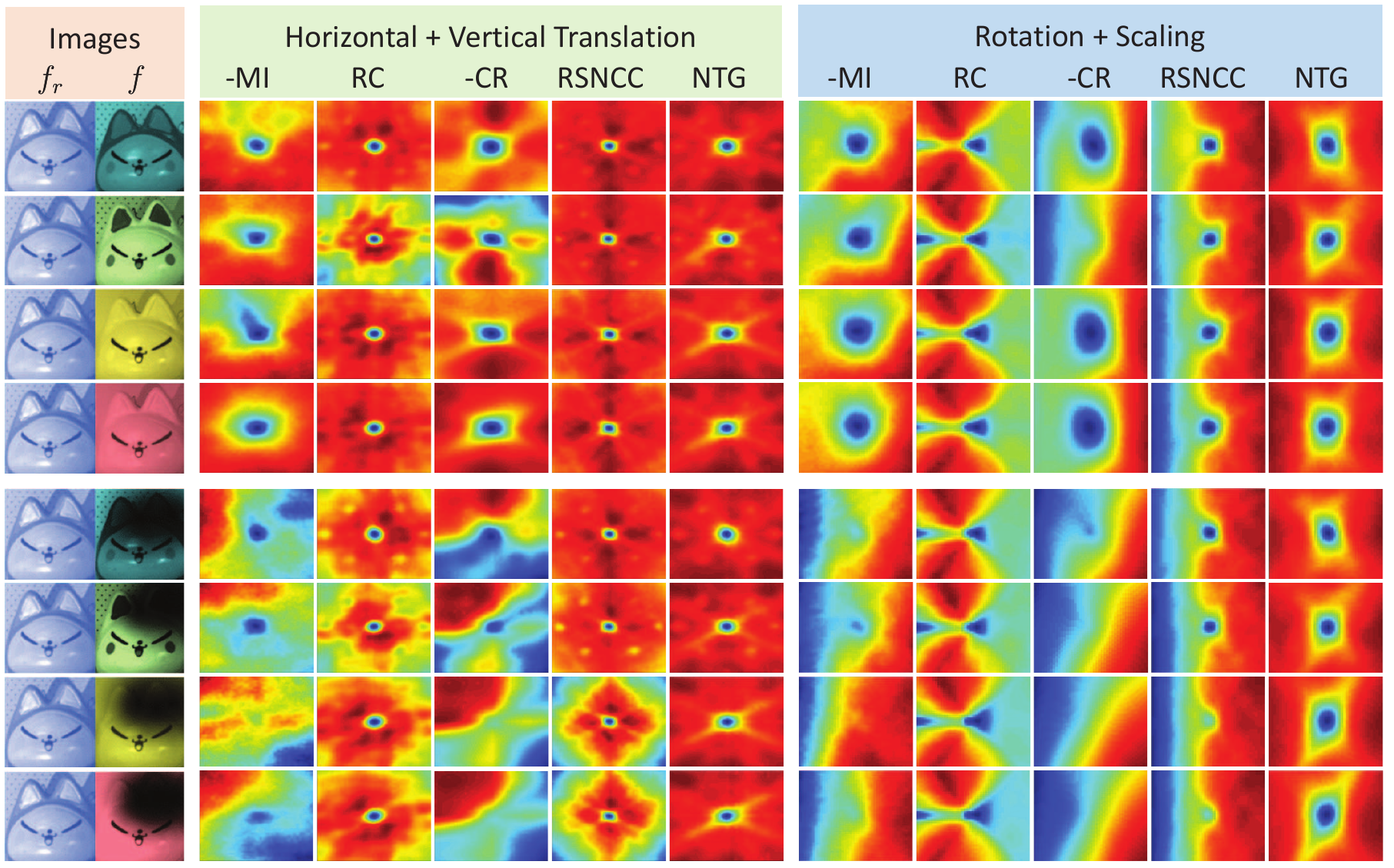}\\
\caption{Effect of local intensity variation on different measures. Image of channel $1$ ($400$ nm) is the reference image and images of channel $5$ ($480$ nm), $9$ ($560$ nm), $12$ ($620$ nm), $16$ ($700$ nm) are the floating images. Cost maps of negative MI, RC, negative CR, RSNCC, and NTG with respect to horizontal and vertical translations ( in range $[-20,20]$ pixels) are shown from column $3$ to $7$. Cost maps with respect to rotation (in range $[-10^{\circ},10^{\circ}]$) and scaling (in range $[0.8,1.2]$) are shown from column $8$ to $12$. The centers of the maps correspond to the ground truth parameters, i.e. $(0,0)$ for translations and $(0,1)$ for rotation and scaling. In the color maps, lower values are in blue and higher values are in red.}\label{fig:localvariation}
\end{figure*}

Figure~\ref{fig:localvariation} shows the effect of local intensity variation on different measures. In the experiment, channel $1$ ($400$ nm) was used as the reference and channels $5$ ($480$ nm), $9$ ($560$ nm), $12$ ($620$ nm), $16$ ($700$ nm) were used as the floating ones. The synthetic affine transforms include horizonal and vertical translations, as well as rotation and scaling. The cost function maps of negative MI, negative CR, RSNCC and NTG are shown in Fig.~\ref{fig:localvariation} from column $3$ to $12$. The top half of the figure shows the cost maps of different measures in case of rapid local intensity variation (see the spots on cat's face and ears). The bottom half shows the cost maps in case of additional slow local intensity variation. The slow variation was simulated by dark clouds whose intensity gradually decreases. The centers of the maps correspond to the ground truth parameters, i.e. $(0,0)$ for the $(x,y)$ translation and $(0,1)$ for the rotation and scaling. In these cost maps, lower cost values are encoded in blue and higher cost values are encoded in red. The locations of the global minimal costs correspond to the estimated transform parameters.

It is observed that MI, RC, CR, and RSNCC are more or less influenced by intensity variation. The reason is that the assumptions like functional mapping, unimodality, and local linearity are violated when local intensity varies. On the contrary, NTG exhibits an excellent performance. The cost function of NTG is scarcely influenced by the varying local intensity. Its robustness to intensity variation takes advantages from the combination of the gradient operation and $L_1$ norm sparseness measure.

\subsection{Registration Accuracy}
In \cite{brauers2011} a pattern board was used to correct the geometric distortion in a multispectral imaging system. Instead, in this work, we employed the pattern board shown in Fig. \ref{fig:database} to evaluate the registration measures. In the imaging procedure, we kept the imaging conditions (e.g., imaging distance and focal length) fixed, but changed the objects to be imaged. The $10$ different scenes are illustrated in Fig. \ref{fig:database}. Note that, as the imaging conditions were fixed, the transform parameters of any individual channel with respect to the reference channel (No. $9$) should be identical in all the captured scenes. In the experiment, we computed the rigid transforms from the sample images (\emph{S1}, \emph{S2}, $\cdots$, \emph{S10}), and the obtained transforms were then applied on the pattern image. For each channel of the pattern image, the displacements between the circle centers on the registered channel image and those on the reference channel image were computed. Finally, the root mean square error (RMSE) of all these displacements was computed as the registration error of each channel.

Table~\ref{tab:RMSE} lists the median RMSE (in pixels) of the $10$ samples produced by different registration measures. It is observed that NTG yields subpixel accuracy and performs comparative to or better than its competitors.

\begin{figure}[!h]
  \centering
  \includegraphics[scale=0.90]{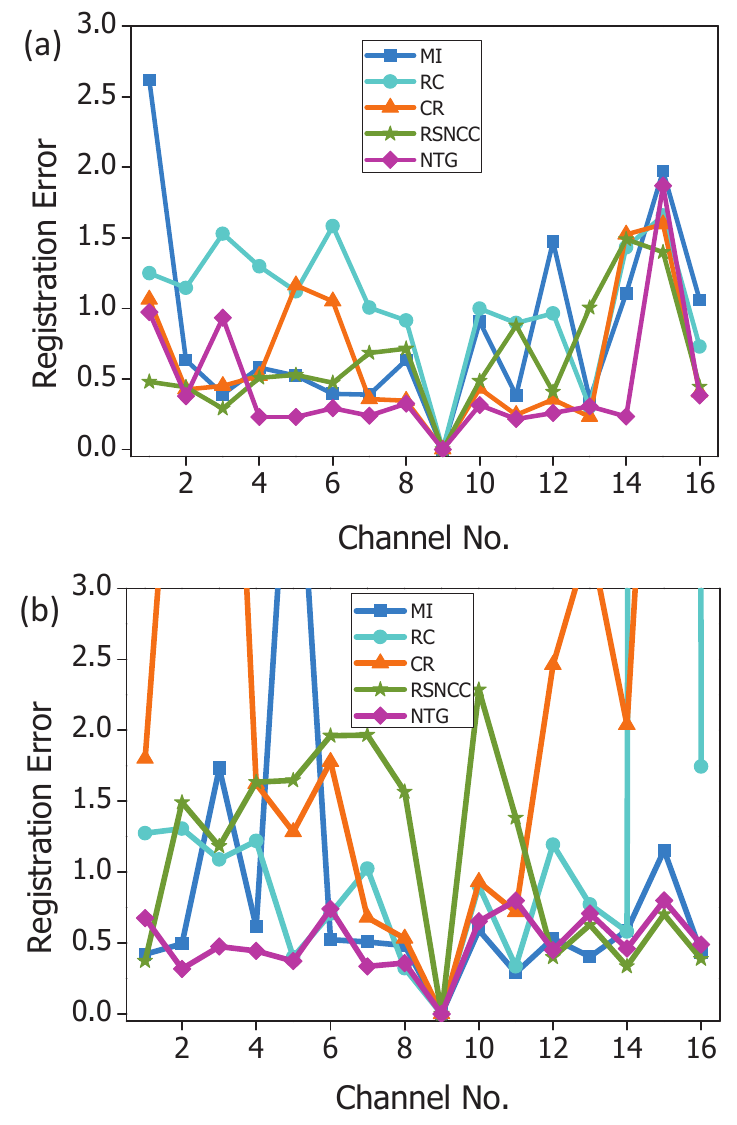}\\
  \caption{Registration errors (in pixels) produced by different measures. (a) Sample \emph{S1}. (b) Sample \emph{S2}. }\label{fig:errorcurve}
\end{figure}

\begin{figure*}[tb]
  \centering
  \includegraphics[scale=0.95]{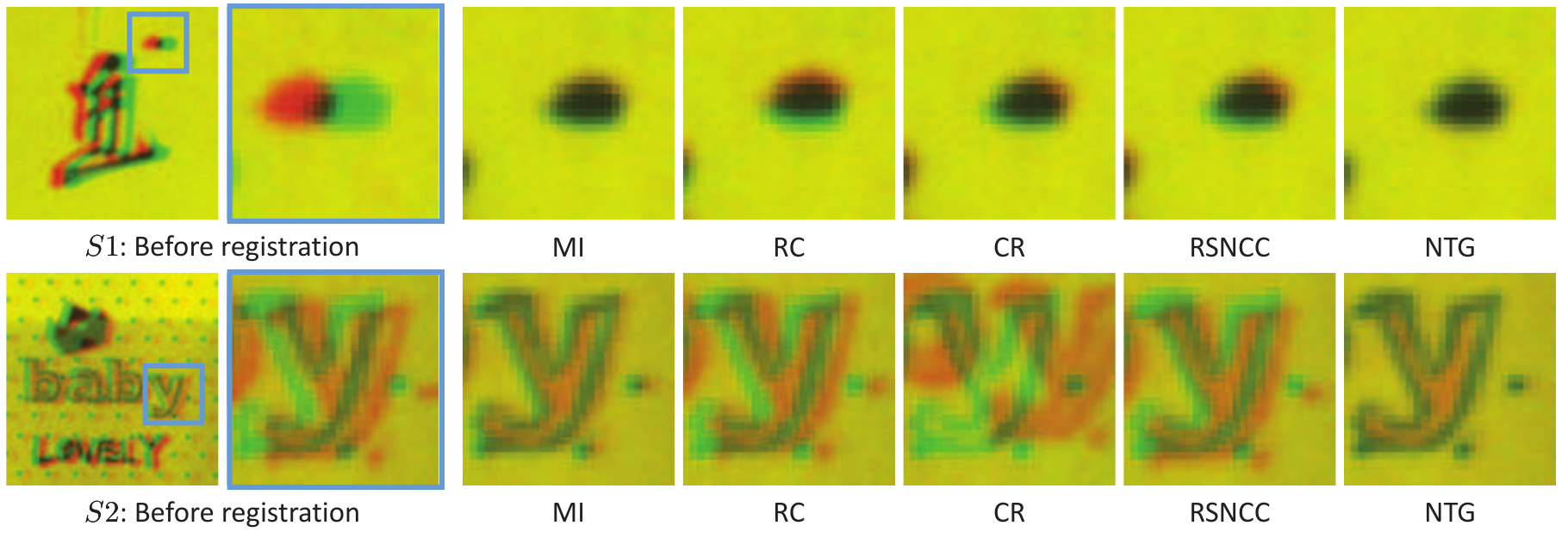}\\
  \caption{Close-up views of multispectral image registration results of samples \emph{S1} and \emph{S2} using different measures. The two channel images, which are in pseudocolor, are displayed in an overlapping manner. }\label{fig:comparison}
\end{figure*}

As an example, Fig.~\ref{fig:errorcurve} shows the distributions of registration errors of the samples \emph{S1} and \emph{S2} in different channels. It is observed that the registration errors by the proposed method are generally smaller than those by other methods. Figure \ref{fig:comparison} further shows the close-up views of image registration results of the samples \emph{S1} and \emph{S2}. Here channel $2$ was chosen to be the floating image and channel $9$ was set as the reference image. The two channel images before and after registration are displayed in an overlapping manner with pseudocolors. It is observed that, for the sample \emph{S1}, small misalignment exists in the registration results by MI, RC, CR, and RSNCC. Rather severe misalignments take place in the registration results of sample \emph{S2}. In comparison, the floating image is robustly and accurately registered by the proposed NTG measure.

\begin{table*}
\renewcommand
\arraystretch{1.5}
\centering
\caption{Registration errors produced by different measures (in pixels). The smallest errors are in bold.}\label{tab:RMSE}
\begin{tabular}{c|c|c|c|c|c|c|c|c|c|c|c|c|c|c|c|c}
  \hline\hline
  Channel No.   & 1 & 2 & 3 & 4 & 5 & 6 & 7 & 8 & 9 & 10 & 11 & 12 & 13 & 14 & 15 & 16 \\\hline
  MI        & 1.15 & 0.68 & 0.66 & 0.61 & 0.61 & 0.51 & 0.58 & 0.56 & ~-~ & 0.68 & 0.58 & 0.66 & 0.61 & 0.85 & 1.47 & 1.27 \\\hline
  RC        & 2.12 & 1.34 & 1.24 & 1.26 & 1.17 & 0.75 & 0.52 & 0.58 & ~-~ & 0.99 & 0.81 & 0.94 & 1.23 & 0.81 & 1.05 & 1.02 \\\hline
  CR        & 3.50 & 1.81 & 1.95 & 1.47 & 1.22 & 0.82 & 0.62 & 0.48 & ~-~ & 0.79 & 0.60 & 0.98 & 0.82 & 1.20 & 2.12 & 2.98 \\\hline
  RSNCC     & \textbf{0.38} & 0.90 & 0.92 & 1.28 & 1.49 & 1.65 & 1.89 & 1.89 & ~-~ & 1.74 & 1.20 & 0.85 & 1.30 & 0.90 & 1.56 & \textbf{0.43} \\\hline
  Proposed NTG       & 0.80 &\textbf{0.37} &\textbf{0.43} &\textbf{0.32} &\textbf{0.26} &\textbf{0.35} &\textbf{0.23} &\textbf{0.24} &~-~ &\textbf{0.36} & \textbf{0.32} &\textbf{0.33} &\textbf{0.34} &\textbf{0.27} &\textbf{0.95} &0.45 \\\hline
  \hline
\end{tabular}
\end{table*}

\begin{figure*}[htb]
  \centering
  \includegraphics[scale=0.8]{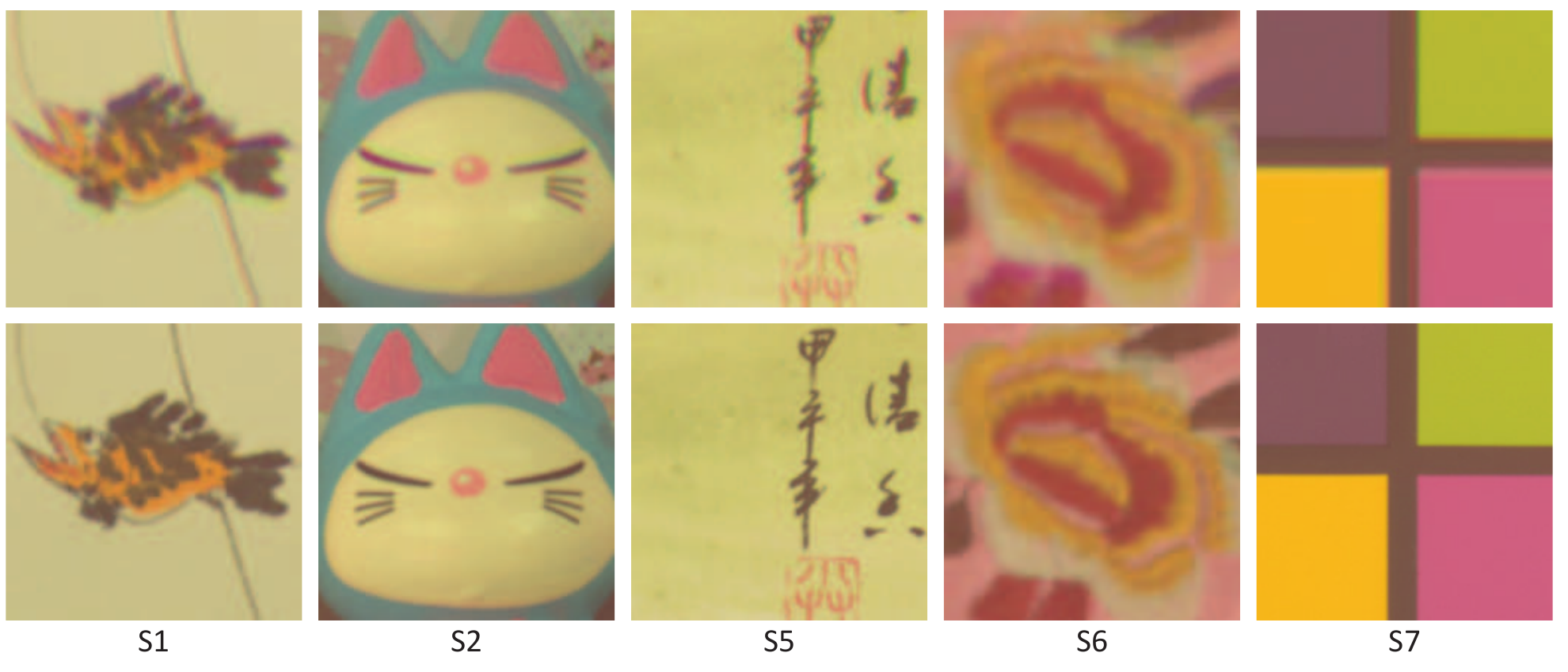}\\
  \caption{Examples of multispectral image registration using NTG. The first and second rows show multispectral images before and after registration. All multispectral images are displayed in RGB. }\label{fig:multispectral}
\end{figure*}

\section{Applications}\label{sec:app}
It is mentioned in Section \ref{sec:intro} that the acquired multispectral images suffer from misalignment and out-of-focus blur. We discuss the application of the proposed method in two circumstances. In the first circumstance (Subsection \emph{A}), the multispectral image has a focusing device \cite{shen2012autofocus} and hence the acquired multispectral images are well focused. Thus the proposed registration method can be directly applied on the sharp multispectral images. In the second circumstance (Subsection \emph{B}), as no focusing device is employed, we must deal with both misalignment and out-of-blur. We will further show in Subsection \emph{C} that, in additional to multispectral image, the proposed method can also be applied on multimodal and other images.

\subsection{Multispectral Image Registration}
Figure \ref{fig:multispectral} shows some registration results of well-focused multispectral images of real scenes using the proposed method. The chromatic abberations caused by image misalignment are obvious in the original acquired multispectral images. By applying image registration, the multispectral images are well aligned and accordingly the chromatic abberations disappear.

\subsection{Multispectral Image Restoration}
\begin{figure}[tb]
  \centering
  \includegraphics[scale=0.66]{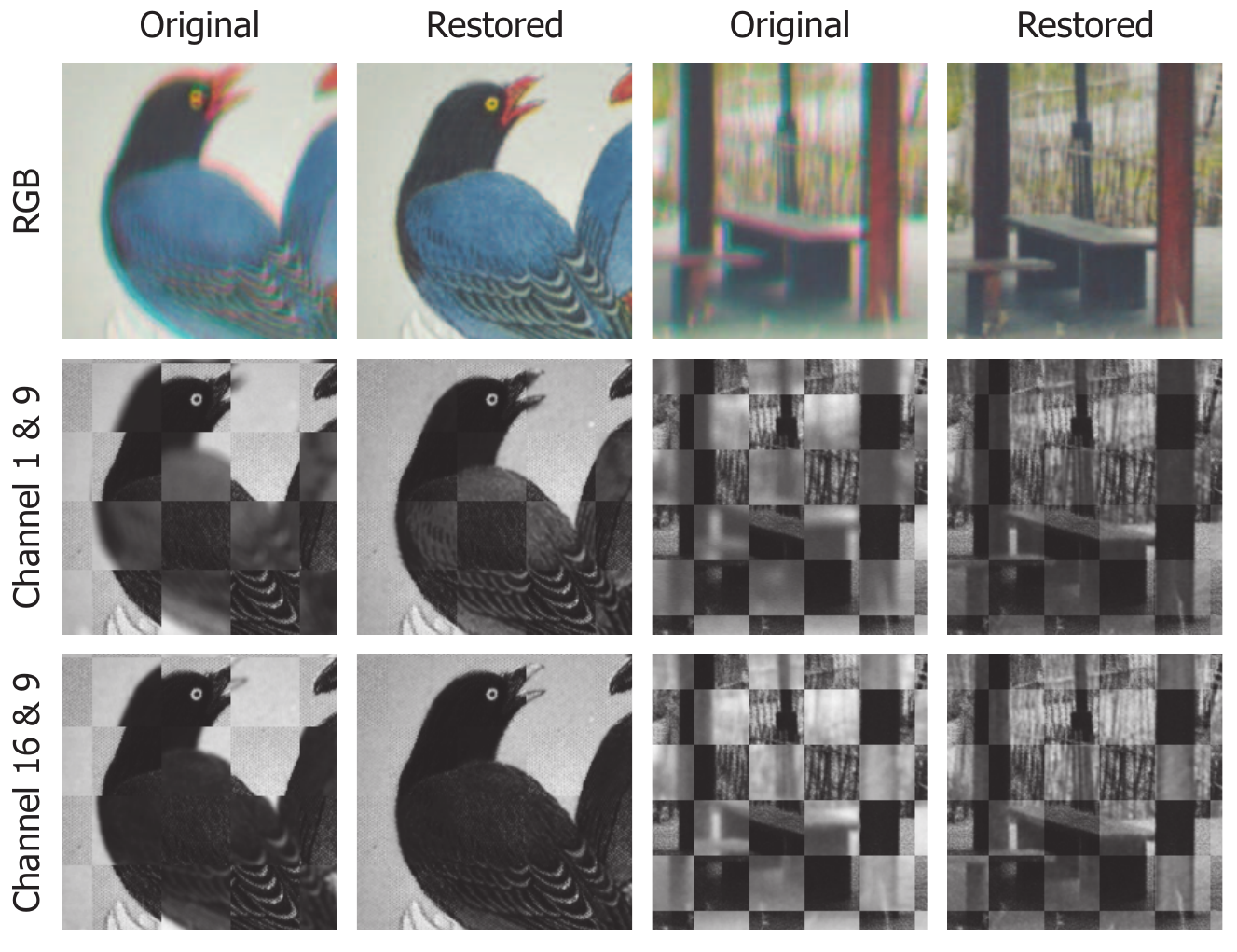}\\
  \caption{Multispectral image restoration (registration and deblurring) results. The quality of the multispectral images (displayed in RGB) is much improved after image restoration. The two blurriest channel images (Nos. $1$ and $16$) are displayed in composite view with respect to the reference channel image (No. $9$).  }\label{fig:deblurring}
\end{figure}

In multispectral restoration, our aim is to remove both misalignment and out-of-focus blur. The process is as follows. First, each channel images were initially deblurred using a single-image-based blind deconvolution method \cite{oliveira2014parametric}. Second, the proposed method was applied to register the deblurred individual channel images with respect to the reference channel image. Third, the state-of-the-art interchannel-correlation-based deblurring method \cite{chen2015multispectral} was applied on the registered blurry images, and the high-quality deblurred multispectral image was finally obtained. Figure~\ref{fig:deblurring} shows the multispectral image restoration results using the above process. For illustration, the two blurriest channels (Nos. $1$ and $16$) and the reference channel (No. $9$) are displayed in composite views. It is observed that quality of the multispectral image (displayed in RGB) are much improved after image restoration. From the composite views in the second and third rows, image registration and deblurring have been successfully accomplished in the $1$th and $9$th channel images.

\begin{figure*}[tb]
  \centering
  \includegraphics[scale=1]{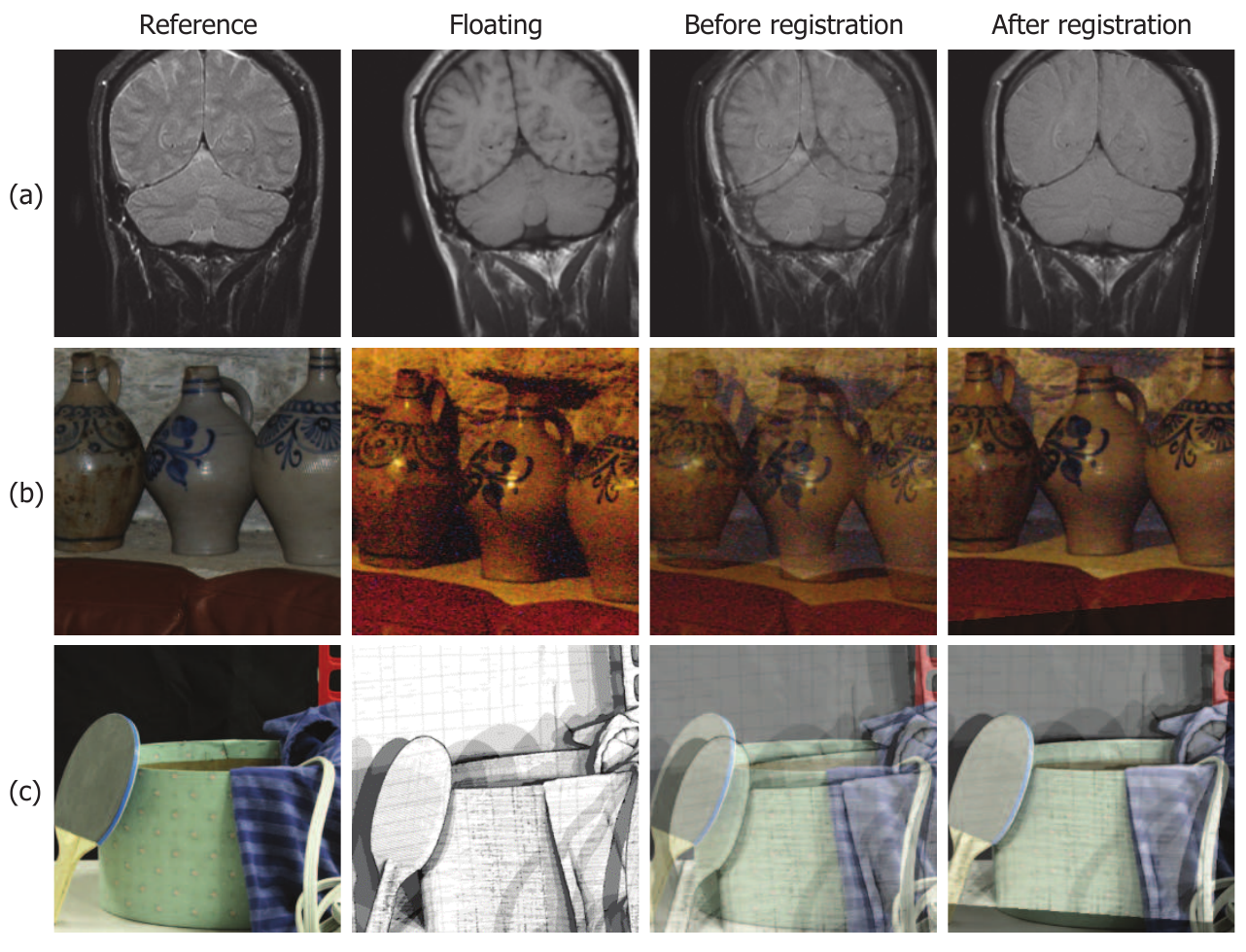}\\
  \caption{Other registration tasks using NTG. (a) Medical images. (b) Flash/no-flash images. (c) RGB and disparity images.  }\label{fig:application}
\end{figure*}

\subsection{Other Registration Tasks}
The proposed method is also applicable to a variety of registration tasks. We present some typical applications in image processing and computer vision. Figure \ref{fig:application} shows the registration results on medical images, flash/no-flash images \cite{petschnigg2004digital}, RGB/disparity images\cite{scharstein2014high}. In these image pairs, both rotation and translation have been simulated in the floating images. It is observed that, despite the rapidly/slowly intensity variations, the proposed method produces quite satisfactory registration results.

\section{Conclusions}\label{sec:conclusion}
 This paper proposed a new measure, namely normalized total gradient (NTG), for multispectral image registration. The employment of NTG was based on the observation that the gradient of difference image becomes sparser when the two channel images are well aligned. An registration framework, which consists of image pyramid and global/local optimization, was introduced for general affine transform. Experimental results showed that the proposed NTG measure is well suited to multispectral image registration and outperforms the conventional measures. It was validated that the proposed registration method can be applied in multispectral image restoration (registration and deblurring) and other relevant registration tasks.

A limitation of this work is that the proposed method is only implemented for rigid (affine) image registration, which is only applicable to static and rigid objects. However, nonrigid registration is required when objects are deformable or moving in the image acquisition process. Hence, in the future we plan to incorporate the NTG measure into nonrigid registration methods to further extend its applications.

\bibliographystyle{IEEEtran}
\bibliography{REF}

\end{document}